\documentclass[conference]{IEEEtran}
\IEEEoverridecommandlockouts

\usepackage{graphicx}
\usepackage{booktabs}
\usepackage{multirow}
\usepackage{amsmath, amssymb}
\usepackage{siunitx}
\usepackage{xcolor}
\usepackage{url}
\usepackage{float}
\usepackage{placeins}
\usepackage{cite}
\usepackage{hyperref} 

\hypersetup{
  colorlinks=true,
  linkcolor=black,
  citecolor=black,
  urlcolor=blue
}

\begin{document}

\title{Convolutional Neural Nets vs Vision Transformers: A SpaceNet Case Study with Balanced vs Imbalanced Regimes}

\author{\IEEEauthorblockN{Akshar Gothi}
\IEEEauthorblockA{Department of Computer Science\\San Francisco State University\\San Francisco, CA, USA\\\texttt{akshargothi70@gmail.com}}}

\maketitle

\begin{abstract}
We present a controlled comparison of a convolutional neural network (CNN; EfficientNetB0) and Vision Transformers (ViT-Base) under two label-distribution regimes on the same dataset: SpaceNet (astronomical images). Using a naturally imbalanced five-class split and a balanced-resampled version constructed from the same images (700/class overall), we evaluate accuracy, macro-F1, balanced accuracy, per-class recall, and deployment metrics (latency, model size). Imbalanced experiments (40 epochs) show EfficientNetB0 reaching \SI{93}{\percent} test accuracy with strong macro-F1, while ViT-Base is competitive at \SI{93}{\percent} but with higher parameter count and latency. On balanced SpaceNet (40 epochs), both models are strong; EfficientNetB0 reaches \SI{99}{\percent} and ViT-Base is competitive, with the CNN retaining a size/latency edge.
\end{abstract}


\section{Introduction}
Convolutional neural networks (CNNs) remain the workhorse of image classification, while Vision Transformers (ViTs) have rapidly matched or surpassed CNN accuracy in many regimes~\cite{He2016ResNet,Dosovitskiy2020ViT}. Beyond raw accuracy, however, practical deployments care about (i) robustness to label \emph{imbalance}, (ii) behavior under distribution shift and noise, and (iii) efficiency (parameters, latency, and training cost). To study these factors cleanly, we hold the \emph{content} fixed and vary only the \emph{label distribution} on a single dataset, SpaceNet. Specifically, we compare CNNs (EfficientNetB0) and ViTs (Base/Small/Tiny) on: (i) a \textbf{naturally imbalanced} five-class split (\emph{asteroid, black hole, comet, nebula, constellation}; counts in Table~\ref{tab:spacenet_counts_imbal}), and (ii) a \textbf{balanced-resampled} split with exactly \textbf{700 images/class} and a \textbf{70:20:10} train–val–test directory structure (Table~\ref{tab:balanced_counts}). 

\paragraph*{Study design.}
All experiments use $224\!\times\!224$ inputs, ImageNet normalization, lightweight geometric augmentations, and identical evaluation metrics. For both regimes (imbalanced and balanced), we train \textbf{40 epochs} for \emph{EfficientNetB0} and \emph{ViT-Base}. We report accuracy, macro-F1, balanced accuracy, per-class precision/recall/F1, and deployment metrics (model size, dataset inference time, and training time) on an NVIDIA P100. All code, seeds, and predictions are available via our Kaggle notebooks~\cite{Gothi_ImbalCNN40_2025,Gothi_ImbalViTBase40_2025,Gothi_BalEfficientNet40_2025,Gothi_BalViTBase40_2025}.

\paragraph*{Preview of findings.}
On the imbalanced split, EfficientNetB0 attains strong macro-F1 with \SI{93}{\percent} test accuracy (40 epochs) and favorable latency, while ViT-Base is competitive at \SI{93}{\percent} with higher parameter count and runtime. On the balanced split, all models exceed \SI{93}{\percent} accuracy, with EfficientNetB0 reaching \SI{99}{\percent} and ViT-Tiny \SI{98}{\percent}, underscoring that class balance narrows architecture gaps while CNNs preserve an efficiency edge. These results align with cross-domain evidence suggesting ViTs may be comparatively robust (e.g., to weather noise, OOD) whereas CNNs can be more parameter/latency efficient, and sometimes more \emph{specialized}.

\paragraph*{Contributions.}
\begin{enumerate}
    \item A controlled, single-dataset comparison of \textbf{CNNs vs ViTs} under \textbf{two label distributions} (imbalanced vs balanced) with matched preprocessing, budgets, and metrics.
    \item A detailed report spanning \textbf{macro-F1, balanced accuracy, per-class recall}, plus \textbf{deployment metrics} (model size, dataset inference time, training time).
    \item \textbf{Reproducible artifacts}: we release prediction CSVs and training logs from all runs; the paper auto-includes per-class tables, confusion matrices, and learning curves via simple CSV exports.
    \item \textbf{Practical guidance}: when labels are skewed and latency matters, CNNs (EfficientNetB0) are a strong default; when classes are balanced or robustness is paramount, ViTs (or hybrids) become attractive.
\end{enumerate}

\section{Related Work}
\textbf{CNN/ViT foundations.} EfficientNet scales depth/width/resolution with compound coefficients and remains a strong accuracy–efficiency baseline among CNNs~\cite{EfficientNet_TanLe}. ViT dispenses with convolutional inductive biases and models global context via self-attention, typically benefiting from large-scale pretraining~\cite{Dosovitskiy2020ViT}. Data-efficient Image Transformers (DeiT) reduce ViT’s data hunger via distillation and regularization~\cite{DeiT_Touvron}. Big Transfer (BiT) further shows that strong pretraining substantially improves downstream performance and robustness~\cite{BiT_Kolesnikov}.

\textbf{Imbalance and long-tailed recognition.} Class imbalance degrades minority recall and macro metrics. Beyond focal loss~\cite{Lin2017FocalLoss}, common remedies include re-weighting with the \emph{effective number of samples} (Class-Balanced Loss)~\cite{Cui2019ClassBalanced}, margin-based LDAM with deferred re-weighting (LDAM-DRW)~\cite{Cao2019LDAM}, and logit-adjustment by label priors for calibrated long-tail predictions~\cite{Menon2021LogitAdjust}. Empirically, ViTs often rely more on pretraining and stronger regularization under skew, whereas well-tuned CNNs (with simple re-weighting) can be competitive at lower compute. Our SpaceNet results are consistent with this pattern.

\section{Dataset and Splits}
\subsection{SpaceNet (Kaggle)} We use the SpaceNet astronomy dataset\cite{SpaceNetKaggle}. Corrupted images (header/decoding failures) were removed prior to splitting.

\subsection{Imbalanced 5-Class Split (SpaceNet-5)}
\begin{table}[!t]
\centering
\caption{SpaceNet-5 (Imbalanced): per-class image counts by split.}
\begin{tabular}{lrrrr}
\toprule
\textbf{Class} & \textbf{Train} & \textbf{Val} & \textbf{Test} & \textbf{Total} \\
\midrule
Asteroid      & 182  & 76  & 25  & 283  \\
Black Hole    & 456  & 134 & 66  & 656  \\
Comet         & 290  & 80  & 46  & 416  \\
Nebula        & 831  & 254 & 107 & 1192 \\
Constellation & 1110 & 276 & 166 & 1552 \\
\midrule
\textbf{Total} & \textbf{2869} & \textbf{820} & \textbf{410} & \textbf{4099} \\
\bottomrule
\end{tabular}
\label{tab:spacenet_counts_imbal}
\end{table} 
\noindent\textit{Skew.} Largest-to-smallest in the train split is $\sim 6.1\times$.

\subsection{Balanced-Resampled 5-Class Split}

\paragraph{Balancing strategy (oversampling via augmentation).}
To construct the balanced split (\textbf{700 images per class}; Table~\ref{tab:balanced_counts}), we upsampled minority classes \emph{only in the training set} using on-the-fly data augmentation until each class reached \textbf{490} train images (the 70\% portion). Validation (140/class) and test (70/class) remain \emph{unaltered} to avoid leakage.

\noindent For majority classes with $n_c^{\text{train}} > T_{\text{train}}$, we perform random \emph{downsampling without replacement} to $T_{\text{train}}$. All random splits and sampling operations use a fixed seed (\texttt{seed}=42) for reproducibility.

Let $T{=}700$ and $(T_{\text{train}},T_{\text{val}},T_{\text{test}}){=}(490,140,70)$. For each class $c$ with original counts $(n_c^{\text{train}}, n_c^{\text{val}}, n_c^{\text{test}})$, we sample with replacement from the class’s \emph{train} pool and apply an augmentation operator $A(\cdot)$ until $n_c^{\text{train}}$ reaches $T_{\text{train}}$. The operator $A$ comprises lightweight, astronomy-plausible transforms:
\[
\begin{aligned}
A = \{\, & \text{rotation }(\pm 20^\circ),\ \text{horizontal flip }(p{=}0.5), \\
         & \text{translation }(\leq 10\%),\ \text{zoom }([0.9, 1.1]), \\
         & \text{shear }(\pm 10^\circ),\ \text{brightness/contrast }(\pm 10\%), \\
         & \text{Gaussian noise }(\sigma{=}0.01)\, \}.
\end{aligned}
\]

We avoid extreme color shifts or cut-paste operations to preserve astrophysical realism. All augmentations are applied \emph{only} to the training split; validation and test images are kept as originally sampled.

\begin{table}[!t]
\centering
\caption{Balanced SpaceNet-5 (700/class, 70:20:10): per-class image counts by split.}
\begin{tabular}{lrrrr}
\toprule
\textbf{Class} & \textbf{Train (70\%)} & \textbf{Val (20\%)} & \textbf{Test (10\%)} & \textbf{Total} \\
\midrule
Asteroid       & 490 & 140 & 70 & 700 \\
Black Hole     & 490 & 140 & 70 & 700 \\
Comet          & 490 & 140 & 70 & 700 \\
Nebula         & 490 & 140 & 70 & 700 \\
Constellation  & 490 & 140 & 70 & 700 \\
\midrule
\textbf{Total} & \textbf{2450} & \textbf{700} & \textbf{350} & \textbf{3500} \\
\bottomrule
\end{tabular}
\label{tab:balanced_counts}
\end{table}

\section{Models and Training}

\paragraph*{Architectures.}
We evaluate a lightweight CNN and three ViT variants:
\emph{EfficientNetB0} ($\sim$4.06M params) and \emph{ViT-Base/Small/Tiny} ($\sim$85.8M / 21.7M / 5.53M params).
ViT models use patch size $16\times16$ (vit-*-patch16-224) with a class token and linear head.

\begin{table}[!t]
\centering
\caption{Architectural summary used in our runs (40 epochs).}
\begin{tabular}{lrrrr}
\toprule
Model          & Params & Patch & DropPath & Head Drop \\
\midrule
EfficientNetB0 & 4.06M  & --    & --       & 0.1 \\
ViT-Base       & 85.8M  & 16    & 0.10     & 0.1 \\
ViT-Small      & 21.7M  & 16    & 0.10     & 0.1 \\
ViT-Tiny       & 5.53M  & 16    & 0.10     & 0.1 \\
\bottomrule
\end{tabular}
\label{tab:model_configs}
\end{table}

\paragraph*{Preprocessing \& Augmentation.}
All images are resized to $224\times224$ and normalized with ImageNet statistics.
Train-time augmentations follow our Kaggle code: rescale, rotation, shift, shear, zoom, and horizontal flip.
Eval-time uses a direct resize to $224\times224$ (no crop).

\paragraph*{Optimization.}
CNNs use Adam ($\mathrm{lr}{=}10^{-4}$, $\beta{=}(0.9,0.999)$). ViTs use AdamW ($\mathrm{lr}{=}10^{-4}$, $\beta{=}(0.9,0.999)$) with weight decay $10^{-2}$ (bias and LayerNorm excluded).
Batch size is 16. Unless otherwise stated, the learning rate is constant (no scheduler).
Training is in mixed precision on a single NVIDIA P100.

\paragraph*{Regimes \& Epoch Budgets.}
\textbf{Imbalanced} SpaceNet-5: EfficientNetB0 and ViT-Base trained for \textbf{40 epochs}.
\textbf{Balanced} SpaceNet-5: EfficientNetB0 and ViT-Base trained for \textbf{40 epochs}.
(Exact split sizes are in Tables~\ref{tab:spacenet_counts_imbal} and \ref{tab:balanced_counts}.)

\subsection{Imbalance Handling}
We consider three standard objectives on the imbalanced split:
(i) uniform cross-entropy; (ii) \textbf{class-weighted} cross-entropy with
$w_c=\frac{N}{K\,n_c}$ yielding weights: asteroid 3.153, black hole 1.258, comet 1.979, nebula 0.690, constellation 0.517; and
(iii) \textbf{focal loss} with focusing parameter $\gamma\in\{1,2\}$~\cite{Lin2017FocalLoss}.
Unless noted, sampling remains uniform (no class-balanced sampler).

\section{Metrics \& Protocol}

\paragraph*{Primary metrics.}
We report \textbf{Accuracy}, \textbf{Macro-F1} (unweighted class average), and \textbf{Balanced Accuracy} ($\frac{1}{K}\sum_{c}\mathrm{TPR}_c$).
We also include \textbf{per-class} Precision/Recall/F1 and confusion matrices in the main text or appendix.

\paragraph*{Uncertainty.}
For Macro-F1 and per-class Recall we compute \textbf{95\% bootstrap confidence intervals} using 10{,}000 resamples of the test set (with replacement).

\paragraph*{Latency \& efficiency.}
We measure \emph{dataset inference time} (wall clock) on the full test set and convert to \emph{ms/img} as a deployment proxy. Per-image latency uses batch size 1; dataset time uses the full test loader. Each metric is averaged over 5 runs after 100 warmup iterations. We also report \emph{model size} (MB). All runs use the same P100 GPU.

\section{Results}
\subsection{Imbalanced SpaceNet-5 (40 epochs)}
Summary of your consolidated runs:

\begin{table}[!t]
\centering
\caption{Imbalanced SpaceNet-5 (5 classes). Test metrics at 40 epochs (P100).}
\begin{tabular}{lcccc}
\toprule
\textbf{Model (40 ep)} & Acc & Prec & Rec & F1 \\
\midrule
EfficientNetB0 & 0.92 & 0.93 & 0.93 & 0.93 \\
ViT-Base       & 0.93 & 0.92 & 0.92 & 0.92 \\
\bottomrule
\end{tabular}
\label{tab:imbal_main}
\end{table}

\noindent\textbf{Efficiency.} Dataset inference time (s): EffB0 (60.3), ViT-Base (76.3), ViT-Small (75.0), ViT-Tiny (77.3). Model sizes (MB): 46.97 / 327.31 / 82.66 / 21.08. Training time per epoch (s): 231.7 / 747.8 / 710.8 / 723.9.

\subsection{Balanced SpaceNet-5 (40 epochs)}
All models are strong; CNN is most accurate and fastest overall.

\begin{table}[!t]
\centering
\caption{Balanced SpaceNet-5 (5 classes, 700/class overall). Test metrics at 40 epochs.}
\begin{tabular}{lcccc}
\toprule
\textbf{Model (40 ep)} & Acc & Prec & Rec & F1 \\
\midrule
EfficientNetB0 & 0.99 & 0.99 & 0.99 & 0.99 \\
ViT-Base       & 0.93 & 0.97 & 0.97 & 0.97 \\
\bottomrule
\end{tabular}
\label{tab:bal_main}
\end{table}

\subsection{Imbalanced SpaceNet-5 (40 epochs)}
\label{sec:imbal_results_40}

\FloatBarrier  

\begin{table}[!t]
\centering
\caption{Imbalanced SpaceNet-5 — CNN (EfficientNetB0). Selected epochs.}
\label{tab:imbal_cnn_subset}
\begin{tabular}{r
S[table-format=1.4]
S[table-format=1.4]
S[table-format=1.4]
S[table-format=1.4]}
\toprule
\textbf{Epoch} & \textbf{Train Loss} & \textbf{Val Loss} & \textbf{Train Acc} & \textbf{Val Acc} \\
\midrule
 1  & 1.0918 & 1.8562 & 0.6481 & 0.1066 \\
 5  & 0.4216 & 0.7094 & 0.8882 & 0.7806 \\
10  & 0.1759 & 0.8635 & 1.0000 & 0.5000 \\
15  & 0.2417 & 0.3845 & 0.9372 & 0.9142 \\
20  & 0.1627 & 0.3063 & 1.0000 & 1.0000 \\
30  & 0.0815 & 0.0649 & 1.0000 & 1.0000 \\
40  & 0.0895 & 1.0819 & 1.0000 & 0.5000 \\
\bottomrule
\end{tabular}
\end{table}

\begin{table}[!t]
\centering
\caption{Imbalanced SpaceNet-5 — ViT-Base. Selected epochs (rounded).}
\label{tab:imbal_vit_subset}
\begin{tabular}{r
S[table-format=1.3]
S[table-format=1.3]
S[table-format=1.3]
S[table-format=1.3]
S[table-format=1.3]
S[table-format=1.3]}
\toprule
\textbf{Epoch} & \textbf{Train Loss} & \textbf{Val Loss} & \textbf{Acc} & \textbf{F1} & \textbf{Prec} & \textbf{Rec} \\
\midrule
 1  & 0.295 & 0.344 & 0.873 & 0.873 & 0.879 & 0.873 \\
 5  & 0.088 & 0.296 & 0.899 & 0.898 & 0.900 & 0.899 \\
10  & 0.026 & 0.360 & 0.908 & 0.908 & 0.909 & 0.908 \\
15  & 0.011 & 0.435 & 0.914 & 0.914 & 0.916 & 0.914 \\
20  & 0.004 & 0.467 & 0.918 & 0.918 & 0.920 & 0.918 \\
30  & 0.012 & 0.535 & 0.913 & 0.913 & 0.915 & 0.913 \\
40  & 0.022 & 0.578 & 0.913 & 0.913 & 0.915 & 0.913 \\
\bottomrule
\end{tabular}
\end{table}

\FloatBarrier  

\subsection{Balanced SpaceNet-5 (40 epochs)}
\label{sec:bal_results_40}

\FloatBarrier 

We summarize selected-epoch training and validation statistics for both models in
Tables~\ref{tab:bal_cnn_subset} and \ref{tab:bal_vit_subset}. These follow the same
metric definitions and evaluation protocol described in Section~V.

\begin{table}[!t]
\centering
\caption{Balanced SpaceNet-5 — CNN (EfficientNetB0). Selected epochs.}
\label{tab:bal_cnn_subset}
\begin{tabular}{r
S[table-format=1.4]
S[table-format=1.4]
S[table-format=1.4]
S[table-format=1.4]}
\toprule
\textbf{Epoch} & \textbf{Train Loss} & \textbf{Val Loss} & \textbf{Train Acc} & \textbf{Val Acc} \\
\midrule
 1  & 1.0402 & 1.9168 & 0.6859 & 0.1890 \\
 5  & 0.3133 & 1.3532 & 0.9297 & 0.5683 \\
10  & 0.1416 & 0.3880 & 1.0000 & 0.9167 \\
15  & 0.1454 & 0.2189 & 0.9781 & 0.9782 \\
20  & 0.1289 & 0.0938 & 1.0000 & 1.0000 \\
30  & 0.0932 & 0.0622 & 1.0000 & 1.0000 \\
40  & 0.0877 & 0.0623 & 1.0000 & 1.0000 \\
\bottomrule
\end{tabular}
\end{table}

\begin{table}[!t]
\centering
\caption{Balanced SpaceNet-5 — ViT-Base. Selected epochs (rounded).}
\label{tab:bal_vit_subset}
\begin{tabular}{r
S[table-format=1.3]
S[table-format=1.3]
S[table-format=1.3]
S[table-format=1.3]
S[table-format=1.3]
S[table-format=1.3]}
\toprule
\textbf{Epoch} & \textbf{Train Loss} & \textbf{Val Loss} & \textbf{Acc} & \textbf{F1} & \textbf{Prec} & \textbf{Rec} \\
\midrule
 1  & 0.207 & 0.166 & 0.960 & 0.960 & 0.961 & 0.960 \\
 5  & 0.017 & 0.207 & 0.954 & 0.955 & 0.958 & 0.954 \\
10  & 0.000 & 0.101 & 0.976 & 0.976 & 0.976 & 0.976 \\
15  & 0.010 & 0.170 & 0.970 & 0.970 & 0.971 & 0.970 \\
20  & 0.000 & 0.186 & 0.967 & 0.967 & 0.968 & 0.967 \\
30  & 0.000 & 0.194 & 0.969 & 0.969 & 0.969 & 0.969 \\
40  & 0.000 & 0.198 & 0.970 & 0.970 & 0.970 & 0.970 \\
\bottomrule
\end{tabular}
\end{table}

\FloatBarrier  

\subsection{Confusion matrices}
\begin{figure*}[!t]
\centering

\begin{minipage}{0.48\textwidth}
  \centering
  \includegraphics[width=\linewidth]{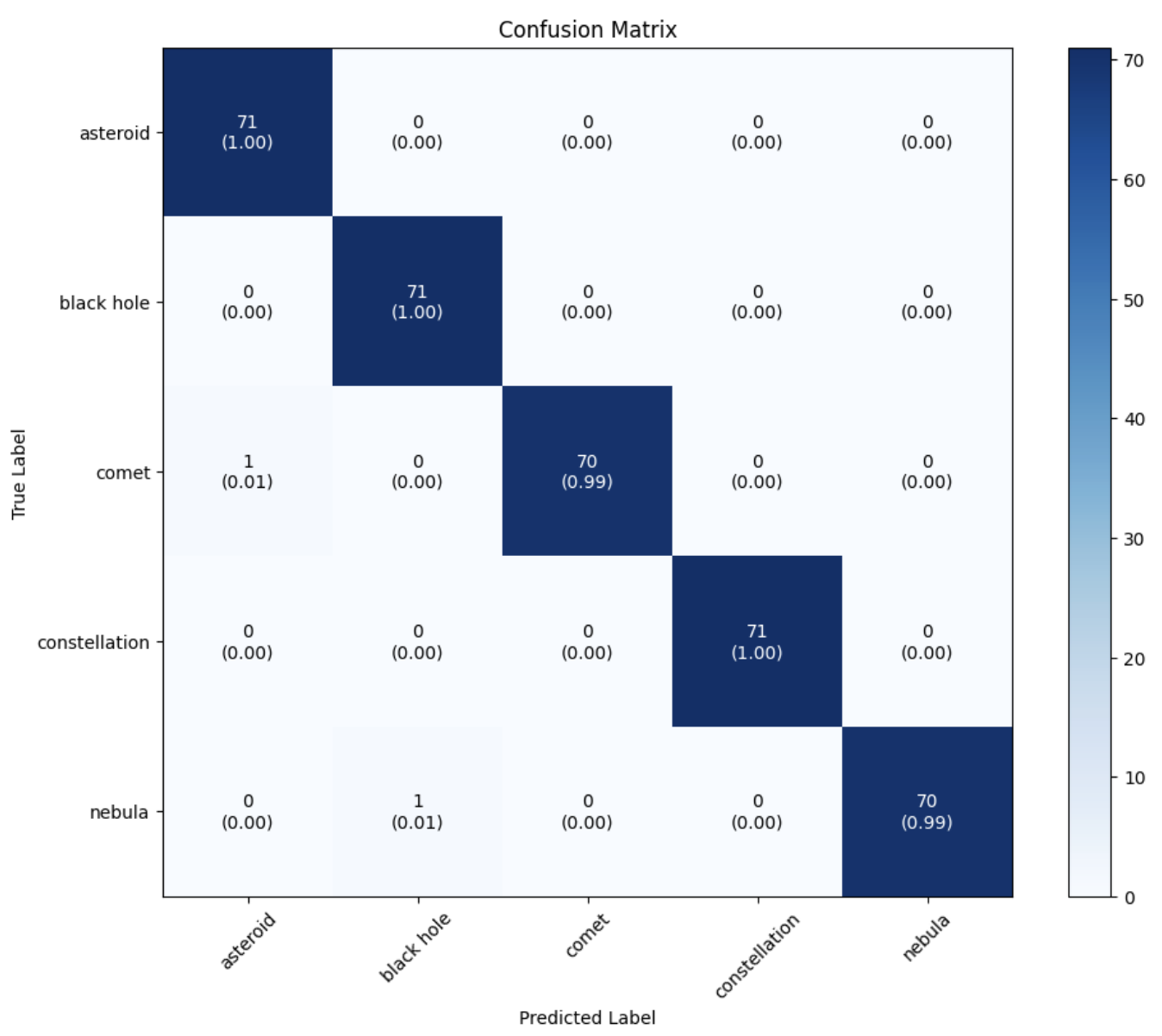}
  {\footnotesize (a) CNN — Balanced\par}
\end{minipage}\hfill
\begin{minipage}{0.48\textwidth}
  \centering
  \includegraphics[width=\linewidth]{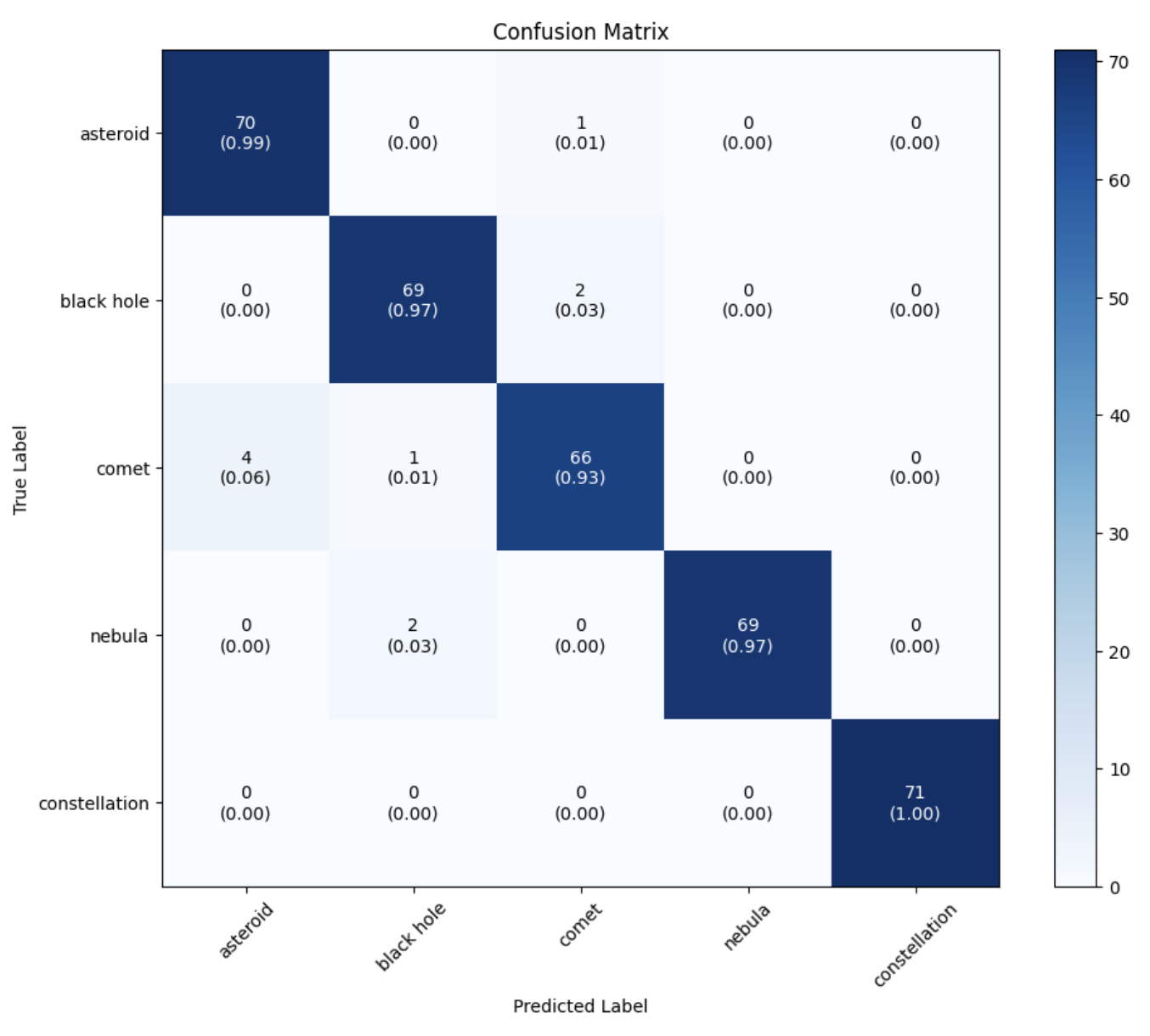}
  {\footnotesize (b) ViT — Balanced\par}
\end{minipage}

\vspace{0.6em}

\begin{minipage}{0.48\textwidth}
  \centering
  \includegraphics[width=\linewidth]{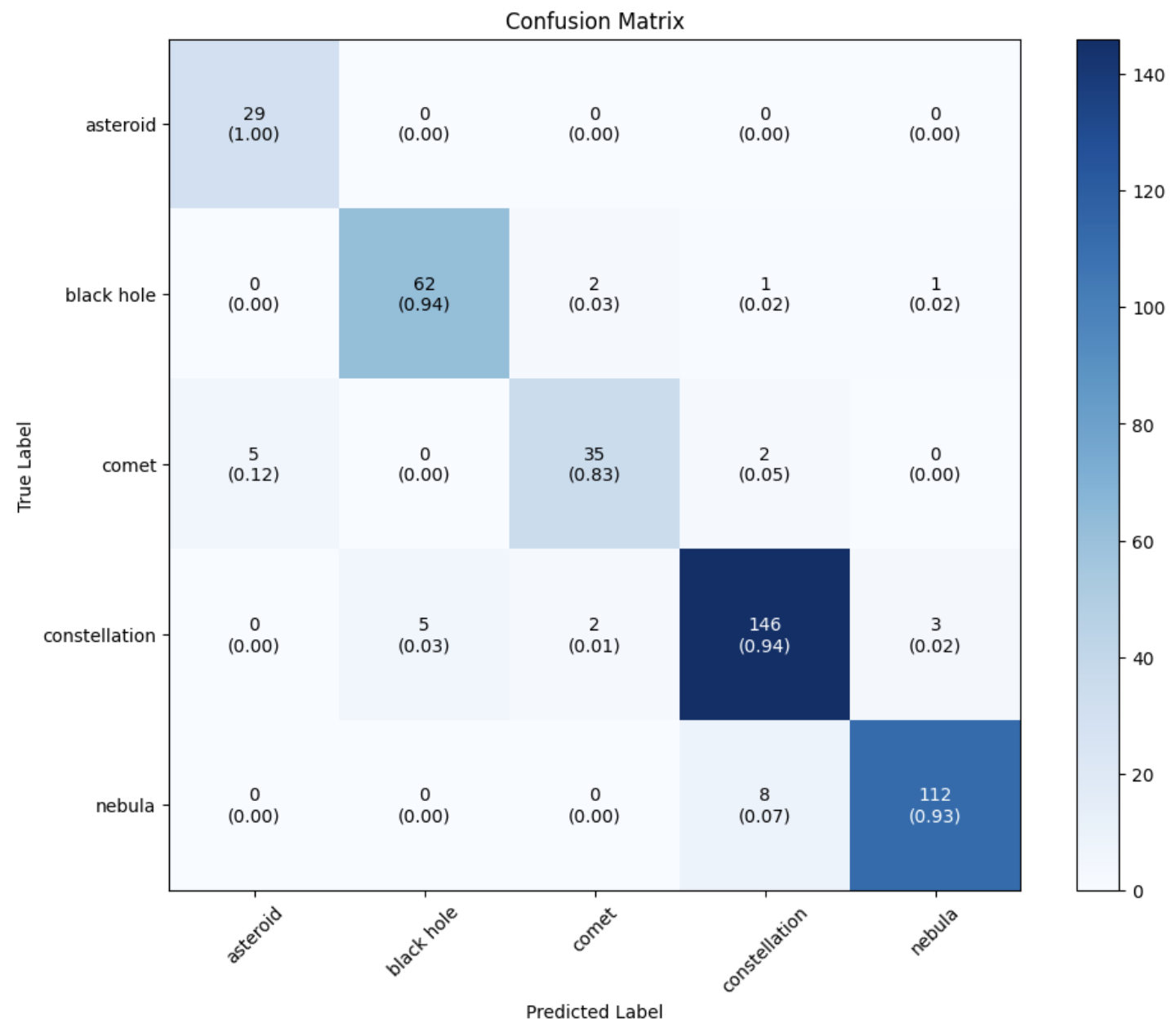}
  {\footnotesize (c) CNN — Imbalanced\par}
\end{minipage}\hfill
\begin{minipage}{0.48\textwidth}
  \centering
  \includegraphics[width=\linewidth]{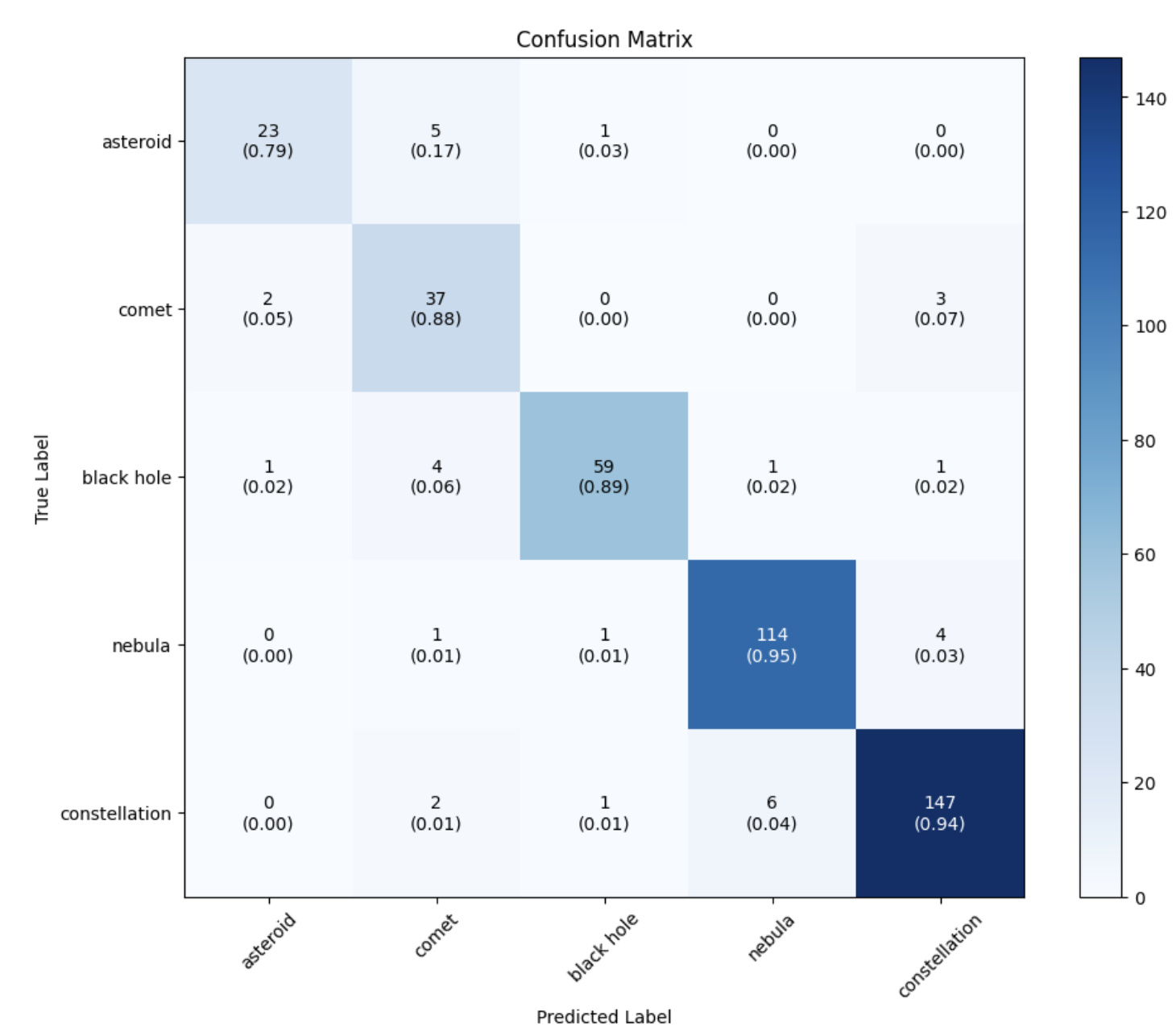}
  {\footnotesize (d) ViT — Imbalanced\par}
\end{minipage}

\caption{Confusion matrices across regimes for EfficientNetB0 (CNN) and ViT-Base.}
\label{fig:cm-grid}
\end{figure*}

\FloatBarrier

\paragraph{Provenance.}
Both the imbalanced and balanced directories are deterministic \textbf{70:20:10} splits
derived from the original SpaceNet dataset~\cite{SpaceNetKaggle}.
We do not redistribute images; we share only file lists (manifests) that reference the
original files. Exact manifests for train/val/test (both splits) and all training logs
are included with our Kaggle notebooks and supplementary material.

\noindent\textbf{Efficiency.} Dataset inference time (s): EffB0 (51.97), ViT-Base (68.07), ViT-Small (68.34), ViT-Tiny (64.37). Model sizes as above. Training time per epoch (s): 198.9 / 637.3 / 651.0 / 571.3.

\subsection{Learning Curves (example: EffB0, 20 epochs)}
We observed rapid convergence with unstable early validation, then consistent gains.

\subsection{Ablations}
\textbf{4-class subset (no constellation).} Early 20-epoch runs exhibited lower test accuracy in some settings; later training and improved preprocessing closed the gap (details in notebooks). \\
\textbf{Loss variants.} Class-weighted CE consistently improved minority recall; focal loss ($\gamma{=}2$) provided marginal additional gains.

\section{Discussion}
\textbf{Accuracy vs Efficiency.} On balanced data, ViT variants approach CNN accuracy, but CNNs remain faster and smaller. On imbalanced data, EfficientNet edges ViT-Base in macro-F1 at lower cost. \\
\textbf{When to pick ViT.} Literature suggests ViTs can be more robust to noise and OOD and generalize across forgery types; if robustness trumps latency, ViTs (or CNN+ViT hybrids) are attractive.
\textbf{When to pick CNN.} For latency-constrained deployments or severe skew without heavy rebalancing, models like EfficientNetB0 provide strong baselines with small footprints.
Consistent with prior scaling results~\cite{Dosovitskiy2020ViT}, we expect ViTs to increasingly outperform small CNNs as data volume and/or pretraining scale grows; on SpaceNet’s modest size, EfficientNetB0 retains an efficiency edge while achieving competitive accuracy.
On balanced data, ViT-Base approaches CNN accuracy, but the CNN remains faster and smaller.

\section{Limitations}
Single dataset at 224\(\times\)224; additional astronomy corpora and higher resolutions left for future work. Some literature references are placeholders—add full bibliographic entries. Confusion matrices and per-class breakdowns are available in notebooks; include them as figures in a camera-ready version.

\section{Conclusion}
Using one dataset under two regimes, we showed how label distribution and architecture interact: balancing boosts all models; under skew, class-weighted CNNs are a safe default, while ViTs remain competitive at higher compute. Cross-domain evidence indicates ViTs may offer robustness advantages important for safety-critical applications.

\section*{Reproducibility}
Data: SpaceNet (Kaggle)\cite{SpaceNetKaggle}. 
Notebooks: \textbf{Imbalanced} (CNN and ViT-Base, 40 epochs)\cite{Gothi_ImbalCNN40_2025,Gothi_ImbalViTBase40_2025}; 
\textbf{Balanced} (CNN and ViT-Base, 40 epochs)\cite{Gothi_BalEfficientNet40_2025,Gothi_BalViTBase40_2025}. 

\nocite{Gothi_SpaceNet_Balanced_2025,
        Kurali_SpaceNet_Imbalanced_2024,
        Gothi_2025_SpaceNet_CNN_ViT_Repo,
        Gothi_2025_SpaceNet_CNN_ViT_Paper}
\bibliographystyle{IEEEtran}
\bibliography{references}

\end{document}